\documentclass[10pt,twocolumn,letterpaper]{article}

\usepackage[margin=0.85in]{geometry}
\usepackage{times}
\usepackage{graphicx}
\usepackage{amsmath,amssymb}
\usepackage{booktabs}
\usepackage{xurl}  
\usepackage[hidelinks]{hyperref}
\usepackage{tikz}
\usetikzlibrary{positioning,arrows.meta,fit,backgrounds,calc}
\usepackage{xcolor}

\title{\Large\bf Modernising the Compressed-Domain Video Captioner:\\
A Controlled Study of SigLIP2 and GPT-2 Substitutions}

\author{
Ashim Nepal\\
Department of Computer Engineering, Pulchowk Campus\\
Institute of Engineering, Tribhuvan University, Nepal\\
{\tt\small official.nepalashim@gmail.com}
\and
Ashok B.K.\\
Department of Computer Engineering, Pulchowk Campus\\
Institute of Engineering, Tribhuvan University, Nepal\\
{\tt\small ashokbk215@gmail.com}
}

\date{}

\begin{document}
\maketitle

\begin{abstract}
Compressed-domain video captioning avoids full video decoding by operating directly on I-frames,
motion vectors and residuals, trading a small amount of accuracy for a large gain in inference
speed. CoCap~\cite{cocap} established this pipeline using a CLIP vision encoder and a shallow
BERT-style multimodal decoder. Both components predate substantially stronger alternatives. We
ask a narrow, controlled question: how much of CoCap's accuracy is limited by these two
components, and which of the two is the binding constraint?

We replace the CLIP I-frame encoder with SigLIP2 and the BERT-style decoder with GPT-2, and
evaluate three configurations (the original pairing, the encoder substitution alone, and both
substitutions together) under identical data, sampling budget and optimisation schedule. All
comparisons are made against our own reproduction of CoCap rather than its published numbers,
because we train on a 4,999-clip subset of VATEX at a reduced sampling budget; absolute values
are therefore not comparable with the original work.

Our reproduction tracks the published result closely: CIDEr and METEOR run slightly above
it (54.9 against 52.7; 23.4 against 23.2), BLEU-4 and ROUGE-L slightly below (29.7 against
31.4; 48.9 against 49.4). We attribute the differences to our evaluation subset rather than
to any improvement in either direction.
We find the two substitutions pull in opposite directions: SigLIP2 alone improves every metric
(+4.5 CIDEr), while adding GPT-2 on top erodes that gain, because a pretrained decoder overfits
4,999 clips within two epochs. We additionally report
inference latency for each configuration, since speed is the property that motivates
compressed-domain captioning in the first place, and an accuracy gain purchased at a latency
cost should be reported as such.
\end{abstract}

\section{Introduction}
\label{sec:intro}

Video captioning systems conventionally decode a video, sample frames, extract 2D and 3D
features, and only then generate text. CoCap~\cite{cocap} removes the decoding and offline
feature-extraction stages entirely by consuming the compressed bitstream: the I-frame supplies
appearance, while motion vectors and residuals, already computed by the video encoder, supply
motion. The result runs roughly $2\times$ faster than the fastest prior end-to-end method while
remaining competitive on accuracy.

That architecture fixes two choices that have since been superseded. The I-frame encoder is
CLIP~\cite{clip}; SigLIP2~\cite{siglip2} has since demonstrated stronger transfer at comparable
cost. The caption decoder is a two-layer BERT-style~\cite{bert} block trained from scratch apart
from its word embeddings, where a pretrained language model is the obvious modern alternative.

Substituting either component is straightforward. What is not obvious is \emph{which}
substitution matters, and whether the answer survives the compressed-domain constraints: a
single I-frame per Group of Pictures, motion vectors at quarter resolution. The visual signal
reaching the decoder here is far sparser than in a frame-based captioner, and it is not
self-evident that a larger decoder can exploit it.

This paper is deliberately narrow in scope. We contribute:

\begin{enumerate}\itemsep2pt
\item A documented reproduction of CoCap on VATEX that tracks the published result across
      metrics (Table~\ref{tab:repro}), together with the source-level fixes the original release requires on a
      current toolchain (Appendix~\ref{sec:repro}).
\item A controlled attribution study over three configurations (CLIP+BERT, SigLIP2+BERT,
      SigLIP2+GPT-2) sharing data, sampling budget, schedule and evaluation, isolating the
      contribution of each substitution.
\item A latency analysis reporting the speed consequence of each substitution alongside its
      accuracy effect.
\item Two empirical observations about the compressed-domain input pipeline that affect anyone
      reproducing this line of work: the GOP sampling budget in the original configuration draws
      a substantial fraction of duplicate GOPs on short clips (Section~\ref{sec:gop}), and half
      the motion-vector channels carry no information on B-frame-free streams
      (Section~\ref{sec:mv}).
\end{enumerate}

We make no claim of improving on CoCap as published. Our training set is roughly one fifth the
size, our evaluation split is a subset, and our sampling budget is reduced; the only valid
comparison is internal, against our own reproduction under identical conditions.

\section{Related Work}
\label{sec:related}

\textbf{Compressed-domain vision.} Operating on the compressed bitstream avoids decoding every
frame and reuses motion information the encoder has already computed. CoViAR~\cite{coviar}
established the approach for action recognition on MPEG-4, back-tracking motion vectors to the
preceding I-frame. MM-ViT~\cite{mmvit} extended it to a multi-modal transformer over I-frames,
motion vectors, residuals and audio. Because MPEG-4 is largely superseded, later work targets
H.264 and H.265, whose more flexible inter-prediction makes the compressed representation harder
to learn from: MVCGC~\cite{mvcgc} learns video representations self-supervised from the mutual
information between RGB frames and motion vectors, and ATTP~\cite{attp} targets real-time
recognition on embedded hardware. CoCap~\cite{cocap} brought the setting to captioning. We work
within its H.264 pipeline unchanged, and our observations in
Sections~\ref{sec:gop} and~\ref{sec:mv} concern that pipeline rather than the model.

\textbf{Video captioning.} One line of work invests in encoder structure over offline-extracted
features: HMN~\cite{hmn} builds a hierarchical modular encoder bridging video and language,
ORG-TRL~\cite{orgtrl} an object relational graph, and SGN~\cite{sgn} groups frames by the word
phrases of a partially decoded caption. A second line removes offline extraction altogether:
SwinBERT~\cite{swinbert} feeds video patches directly to a VidSwin backbone, and
MV-GPT~\cite{mvgpt} pretrains an encoder-decoder end-to-end over frames and transcribed speech.
A third brings in outside knowledge, as TextKG~\cite{textkg} does with knowledge graphs. These
all decode video first. CoCap's contribution is to skip that step, which is what makes its
latency competitive and what makes latency the axis on which any modification to it must be
judged.

\textbf{Pretrained encoders and decoders.} CLIP~\cite{clip} made contrastively pretrained vision
encoders a default initialisation for vision-language tasks, and CoCap adopts it for the
I-frame branch. SigLIP~\cite{siglip} replaced the softmax contrastive objective with a pairwise
sigmoid loss that decouples training from batch size, and SigLIP2~\cite{siglip2} adds
decoder-based pretraining, self-distillation and masked prediction, improving dense and
localisation-sensitive transfer in particular. Those properties matter here, since the action
encoder cross-attends into the I-frame's patch tokens rather than using a pooled vector alone.
On the language side, pairing a pretrained decoder with a visual prefix is well established for
image captioning, ClipCap~\cite{clipcap} being the canonical minimal form. Our results suggest
that the transfer of this recipe to compressed-domain video captioning is bounded by data scale
rather than by architecture.

\section{Method}
\label{sec:method}

\begin{figure*}[t]
\centering
\resizebox{0.95\textwidth}{!}{
\begin{tikzpicture}[
    font=\small,
    box/.style     = {draw, rounded corners=2pt, minimum height=8mm, align=center,
                      inner sep=3pt, fill=gray!8},
    newbox/.style  = {box, fill=orange!22, draw=orange!70!black, very thick},
    inp/.style     = {draw, dashed, rounded corners=2pt, minimum height=7mm, align=center,
                      inner sep=3pt, fill=blue!6},
    feat/.style    = {align=center, inner sep=1pt},
    op/.style      = {draw, circle, inner sep=0.5pt, minimum size=4.2mm},
    ar/.style      = {-{Latex[length=2mm]}, thick},
    lbl/.style     = {font=\scriptsize\itshape, inner sep=1pt},
]

\node[inp] (iframe) {I-frame \\ {\scriptsize $3\times224\times224$}};
\node[inp, below=7mm of iframe] (mv) {Motion vector \\ {\scriptsize $2\times56\times56$}};
\node[inp, below=5mm of mv] (res) {Residual \\ {\scriptsize $3\times224\times224$}};

\node[newbox, right=9mm of iframe, minimum width=26mm] (sig)
    {\textbf{SigLIP2} \\ {\scriptsize vision encoder}};
\node[box, right=9mm of mv, minimum width=26mm] (menc)
    {Motion encoder \\ {\scriptsize ViT, 2 layers}};
\node[box, right=9mm of res, minimum width=26mm] (renc)
    {Residual encoder \\ {\scriptsize ViT, 2 layers}};

\draw[ar] (iframe) -- (sig);
\draw[ar] (mv) -- (menc);
\draw[ar] (res) -- (renc);

\node[op] (plus) at ($(menc.east)!0.5!(renc.east) + (7mm,0)$) {$+$};
\draw[ar] (menc.east) -- ++(3mm,0) |- (plus.west);
\draw[ar] (renc.east) -- ++(3mm,0) |- (plus.west);

\node[box, right=9mm of plus, minimum width=27mm, minimum height=13mm] (act)
    {Action encoder \\ {\scriptsize self-attn $\rightarrow$ cross-attn} \\ {\scriptsize $\rightarrow$ mean}};
\draw[ar] (plus) -- node[lbl, below] {$\mathcal{F}_{\mathrm{BP}}$} (act);

\coordinate (sigout) at ($(sig.east)+(4mm,0)$);
\draw[ar] (sig.east) -- (sigout) |- ([yshift=3mm]act.west)
    node[lbl, pos=0.75, above] {patch tokens};

\node[feat, right=10mm of act] (vis)
    {$\mathcal{V}=[\mathcal{F}^{(n)}_{\mathrm{ctx}},\mathcal{F}^{(n)}_{\mathrm{act}}]$};
\draw[ar] (act) -- node[lbl, above] {$\mathcal{F}_{\mathrm{act}}$} (vis);
\draw[ar] (sigout) to[out=0, in=120] node[lbl, pos=0.4, above] {$\mathcal{F}_{\mathrm{ctx}}$} (vis.north);

\node[box, right=9mm of vis, minimum width=17mm] (proj) {Linear \\ {\scriptsize projection}};
\node[newbox, right=8mm of proj, minimum width=24mm, minimum height=16mm] (gpt)
    {\textbf{GPT-2} \\ {\scriptsize causal decoder} \\ {\scriptsize prefix-LM}};
\node[inp, below=10mm of proj] (cap) {Caption tokens \\ {\scriptsize $\mathcal{T}_{<t}$}};
\node[feat, right=8mm of gpt] (out) {$p(y_t \mid \mathcal{V}, \mathcal{T}_{<t})$};

\draw[ar] (vis) -- (proj);
\draw[ar] (proj) -- (gpt);
\draw[ar] (cap.east) -| ([xshift=-5mm]gpt.south) -- (gpt.south);
\draw[ar] (gpt) -- (out);

\node[lbl, below=3mm of res] (pergop) {$\times M$ per GOP, $\times N$ GOPs};

\begin{scope}[on background layer]
    \node[draw, dashed, rounded corners, fit=(iframe)(res)(act)(vis)(pergop),
          inner sep=4mm,
          label={[font=\scriptsize\bfseries]above:Compressed Video Transformer}] {};
    \node[draw, dashed, rounded corners, fit=(proj)(gpt)(cap)(out),
          inner sep=4mm,
          label={[font=\scriptsize\bfseries]above:Multimodal Decoder}] {};
\end{scope}

\end{tikzpicture}}
\caption{The captioner with both substitutions applied. Highlighted blocks are the components we
replace; the remainder follows CoCap~\cite{cocap}. Per Group of Pictures, one I-frame is encoded
by SigLIP2 and $M$ motion-vector/residual pairs by two lightweight, randomly initialised Vision
Transformers. The action encoder fuses them, using the motion features as queries against the
I-frame's patch tokens. The resulting per-GOP context and action features form a visual prefix
for a causal GPT-2 decoder.}
\label{fig:arch}
\end{figure*}
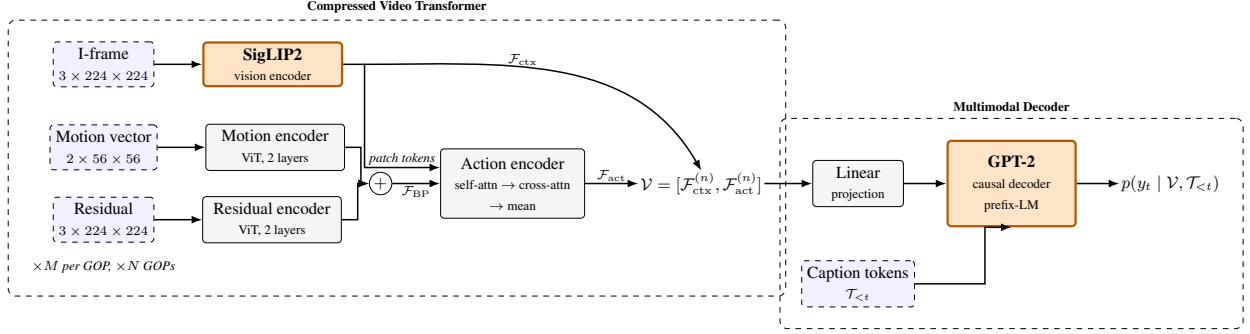

We retain CoCap's structure and change two components. Section~\ref{sec:prelim} summarises what
is inherited; Sections~\ref{sec:siglip} and~\ref{sec:gpt2} describe the substitutions.

\subsection{Preliminaries: the compressed-domain captioner}
\label{sec:prelim}

An H.264 stream is a sequence of Groups of Pictures (GOPs), each beginning with an intra-coded
I-frame $\mathcal{I}_{I}$ followed by inter-coded frames carrying a motion-vector field
$\mathcal{I}_{mv}$ and a residual $\Delta_{res}$. Sampling $N$ GOPs and $M$ inter-coded frames
per GOP, the model encodes
\begin{equation}
\mathcal{F}^{(m,n)}_{\mathrm{BP}} =
  \mathrm{Enc}_{mv}\!\left(\mathcal{I}^{(m,n)}_{mv}\right) +
  \mathrm{Enc}_{res}\!\left(\Delta^{(m,n)}_{res}\right),
\end{equation}
and an action encoder fuses these with the I-frame representation, using the motion features as
queries against the I-frame's spatial tokens so that motion is grounded in appearance. Mean
pooling yields a per-GOP action feature $\mathcal{F}^{(n)}_{\mathrm{act}}$, concatenated with the
per-GOP context feature to form the visual representation
$\mathcal{V} = [\mathcal{F}^{(n)}_{\mathrm{ctx}}, \mathcal{F}^{(n)}_{\mathrm{act}}]$.
The motion and residual encoders are randomly initialised; only the I-frame encoder is
pretrained. We leave all of this unchanged.

\subsection{SigLIP2 I-frame encoder}
\label{sec:siglip}

CLIP's Vision Transformer prepends a \texttt{[CLS]} token whose projected output CoCap uses as
the per-GOP context feature, with the remaining patch tokens serving as cross-attention memory
for the action encoder. SigLIP2~\cite{siglip2} has no \texttt{[CLS]} token: it terminates in a
multi-head attention pooling head in which a single learned probe attends over the patch tokens.
We therefore take the pooled probe output as the context feature and the post-layer-normalised
patch tokens as cross-attention memory, preserving both roles.

At \texttt{patch16-224} resolution the two encoders emit the same number of patch tokens (196),
so the action encoder is structurally unchanged. The representation width moves from CLIP's
512-dimensional projected space to SigLIP2's 768, which propagates to the action encoder and the
decoder interface. Input normalisation changes from ImageNet statistics to SigLIP's
$\mu = \sigma = 0.5$.

\subsection{GPT-2 caption decoder}
\label{sec:gpt2}

CoCap's decoder concatenates the visual tokens with the caption tokens and applies a shifted
causal mask, so that every text position attends to the complete visual representation and to
preceding text only. This is precisely the attention pattern of GPT-2~\cite{gpt2}, which makes
the substitution a prefix-LM rather than a re-architecture: we project the visual tokens into GPT-2's
embedding space, prepend them to the caption's word embeddings, and let the pretrained causal
mask do the rest. A learned type embedding distinguishes context from action tokens and is
zero-initialised, so the head begins as an unmodified GPT-2.

The substitution replaces the CLIP byte-pair vocabulary with GPT-2's. Because GPT-2 has no
dedicated padding token (\texttt{endoftext} serves as both sequence terminator and
pad), padded positions cannot be identified by token identifier, and supervision instead masks
them with $-100$ in the label tensor. SigLIP2-base's width of 768 matches GPT-2's embedding width
exactly, so the visual projection is dimension-preserving.

\section{Experimental Setup}
\label{sec:setup}

\subsection{Data}

We use a 4,999/1,000 clip train/validation subset of VATEX~\cite{vatex}, with ten English
captions per clip, and report BLEU-4, METEOR, ROUGE-L and CIDEr~\cite{cider}, monitoring the last
of these for checkpoint selection. Two clip identifiers in the source annotation map to a single video file with
no recoverable correspondence; both are discarded rather than paired with a video that may not
match. Clips arrive already encoded as H.264 at 240 pixels on the short edge with
\texttt{KeyInt} of 60 and no B-frames, so no re-encoding is performed.

\subsection{Deviations from the published configuration}
\label{sec:deviations}

Table~\ref{tab:deviations} lists every respect in which our setup departs from
CoCap~\cite{cocap}. These are why absolute numbers here cannot be compared with the published
results. Each deviation is applied identically to all three configurations, so the internal
comparison that carries our conclusions is unaffected.

\begin{table}[t]
\centering\small
\begin{tabular}{lrr}
\toprule
 & This work & CoCap~\cite{cocap} \\
\midrule
Training clips             & 4{,}999 & 25{,}991 \\
Evaluation clips           & 1{,}000 & full test split \\
GOPs per clip ($N$)        & 5 & 8 \\
Frames per GOP ($M$)       & 16 & 59 \\
Motion-vector channels     & 2 & 4 \\
Epochs                     & 12 & 20 \\
Batch size                 & 12 & 64 \\
\bottomrule
\end{tabular}
\caption{Deviations from the published configuration, applied uniformly across all three
variants. Sections~\ref{sec:gop} and~\ref{sec:mv} justify the sampling reductions empirically.}
\label{tab:deviations}
\end{table}

\subsubsection{GOP sampling budget}
\label{sec:gop}

CoCap samples $N = 8$ GOPs per clip. VATEX clips are approximately ten seconds, which at
\texttt{KeyInt} of 60 yields a median of five GOPs. Because GOP sampling occurs after the
per-GOP frame sampling, requesting more GOPs than a clip contains yields byte-identical
duplicates. We measure a 35\% duplicate fraction at $N = 8$ and 0\% at $N = 5$. Reducing $N$ to
five therefore costs no information: it removes duplicated visual tokens from the decoder's
input, and the corresponding redundant encoder passes.

\subsubsection{Motion-vector channels}
\label{sec:mv}

The compressed-domain reader returns four motion-vector channels for H.264, laid out as
$[\delta x_{L0}, \delta y_{L0}, \delta x_{L1}, \delta y_{L1}]$, where the $L1$ pair references
future frames and is populated only by B-frames. Our clips contain none: across 20 clips and
5{,}605 motion-carrying frames we observe only I- and P-frames, and the per-channel maximum
absolute values are $[253, 125, 0, 0]$. The $L1$ pair is therefore identically zero and is
discarded, halving the motion tensor at no information cost.

\subsubsection{Frames per GOP}

The reduction from $M = 59$ to $M = 16$ is a genuine loss of per-forward-pass information,
imposed by an 8\,GB memory budget. At $N = 8$, $M = 59$ the residual tensor alone comprises 944
images of $224^2$ per batch; memory occupancy reaches 7{,}710 of 8{,}188\,MiB and throughput
collapses to 0.04 iterations per second. Because the sampler re-draws each epoch, the model
still observes the full GOP over the course of training, but sees only a subset within any
single forward pass.

\subsection{Implementation}

All three configurations share an identical optimisation setup; only the substituted components
differ. We use the Adam variant with decoupled weight decay and gradient clipping at $1.0$ that
CoCap employs, over three parameter groups: randomly initialised parameters (the motion,
residual and action encoders, and the visual projection) at $10^{-4}$; the pretrained vision
backbone at $10^{-6}$; and, where present, the pretrained decoder at $5\times10^{-5}$. Separating
the third group matters, because a rate suited to freshly initialised layers destabilises a
pretrained language model on a training set this small.

The learning rate warms up linearly over the first 10\% of optimiser steps and then decays
per-epoch by $0.91$, reaching $0.32$ of peak by the end of the twelfth epoch. This decay rate is
retuned for the shortened schedule: CoCap's $0.95$ over twenty epochs leaves a twelve-epoch run
at $0.54$ of peak and visibly under-annealed. Training uses label smoothing of $0.1$, a caption
length of 32 tokens, and bfloat16 mixed precision. The optimisation batch is 12, reached as a
per-device batch of 2 with six gradient accumulation steps.

Experiments run on a single RTX 4070 Laptop GPU with 8\,GB of memory. Compressed-domain features
are extracted once ahead of training and cached (20.2\,GB for 6,000 clips, about eight minutes),
since with ten captions per clip a twelve-epoch run would otherwise parse each bitstream around
120 times. An epoch takes approximately 1\,h\,12\,m to 1\,h\,19\,m for the configurations
using the shallow decoder and 1\,h\,28\,m for the GPT-2 configuration, whose twelve decoder
layers slow both the training step and the greedy validation decode.

\section{Results}
\label{sec:results}

\subsection{Reproduction fidelity}

\begin{table}[t]
\centering\small
\begin{tabular}{lrrrr}
\toprule
& B4 & M & R & C \\
\midrule
CoCap, published~\cite{cocap} & 31.4 & 23.2 & 49.4 & 52.7 \\
Our reproduction              & 29.7 & 23.4 & 48.9 & 54.9 \\
\bottomrule
\end{tabular}
\caption{Our CoCap reproduction against the published VATEX result. The evaluation splits differ
(Section~\ref{sec:deviations}), so this indicates fidelity rather than being a like-for-like
comparison. Two of the four metrics run slightly above the published values and two slightly
below, despite one fifth the training data; we attribute the differences to the evaluation
subset rather than to any improvement.}
\label{tab:repro}
\end{table}

The reproduction converges rather than being truncated: per-epoch CIDEr gains fall from $+2.5$ at
epoch~7 to $+1.2$, $+0.7$, $+0.1$ and $+0.2$ over the four epochs that follow, so the
twelve-epoch budget is
sufficient. We note in passing that the individual metrics peak at different epochs (BLEU-4 at
epoch~8, ROUGE-L at epoch~9, CIDEr at epoch~11), so the choice of reported checkpoint depends on
the monitored metric. We monitor CIDEr throughout, for all configurations.

\subsection{Attribution}

\begin{table}[t]
\centering\small
\begin{tabular}{llrrrr}
\toprule
Encoder & Decoder & B4 & M & R & C \\
\midrule
CLIP    & BERT-style & 29.67 & 23.40 & 48.92 & 54.85 \\
\textbf{SigLIP2} & \textbf{BERT-style} & \textbf{31.01} & \textbf{24.06} & \textbf{49.43} & \textbf{59.33} \\
SigLIP2 & GPT-2 & 29.71 & 23.63 & 48.75 & 56.56 \\
\bottomrule
\end{tabular}
\caption{Attribution over the two substitutions, under identical data, sampling budget and
schedule. Each row reports the epoch of highest validation CIDEr. Single seed throughout; see
Section~\ref{sec:limitations}.}
\label{tab:main}
\end{table}

Substituting the encoder alone accounts for the entire improvement: SigLIP2 paired with the
original shallow decoder gains $+4.48$ CIDEr over the reproduction, and improves every other
metric as well ($+1.34$ BLEU-4, $+0.66$ METEOR, $+0.51$ ROUGE-L). Adding the GPT-2 decoder on
top does not compound that gain; it erodes it, to $+1.71$ CIDEr.

The reason lies in the training dynamics rather than the peak scores, and
Table~\ref{tab:stability} makes it explicit. Both configurations using the shallow decoder train
to convergence and hold their best score to the end. The GPT-2 configuration instead peaks at
epoch~2 and declines monotonically for the following nine epochs, finishing $11.3$ CIDEr below
its own peak and $9.6$ below the reproduction baseline, while its training loss falls by a factor
of $5.7$ (from $89.8$ to $15.6$). A 124M-parameter pretrained decoder memorises 4,999 clips; the
two-layer decoder trained from scratch lacks the capacity to do so, which at this data scale is
a virtue rather than a limitation.

\begin{table}[t]
\centering\small
\begin{tabular}{llrrr}
\toprule
Encoder & Decoder & Best C & Epoch & Final C \\
\midrule
CLIP    & BERT-style & 54.85 & 11 & 54.85 \\
SigLIP2 & BERT-style & 59.33 & 9 & 59.11 \\
SigLIP2 & GPT-2 & 56.56 & 2 & 45.30 \\
\bottomrule
\end{tabular}
\caption{Training stability. The shallow decoder holds its peak through to the end of training;
the pretrained decoder reaches its peak at epoch~2 and loses 11.3 CIDEr over the nine epochs
that follow.}
\label{tab:stability}
\end{table}

\subsection{Inference latency}

\begin{table}[t]
\centering\small
\begin{tabular}{llrrr}
\toprule
Encoder & Decoder & Visual & Decode & Total \\
\midrule
CLIP    & BERT-style & 50.6 & 184.3 & 249.2 \\
SigLIP2 & BERT-style & 50.6 & 173.4 & \textbf{222.9} \\
SigLIP2 & GPT-2 & 55.0 & 400.1 & 458.7 \\
\bottomrule
\end{tabular}
\caption{Per-clip inference latency in milliseconds (median of 50 runs, batch size one, RTX 4070
Laptop), measured on the same hardware as the accuracy results. Visual encoding and caption
decoding are reported separately because the two substitutions affect them differently.}
\label{tab:latency}
\end{table}

The encoder substitution is free. SigLIP2 and CLIP ViT-B/16 share depth, width and patch size, so
their visual passes are indistinguishable at 50.6\,ms each; the small difference in total latency
($222.9$ against $249.2$\,ms) lies within the run-to-run spread of these measurements
($\sigma \approx 30$--$36$\,ms) and we do not claim it as a speedup. The defensible statement is
that SigLIP2 delivers $+4.48$ CIDEr at no measurable latency cost, which matters for a method
whose premise is inference speed.

The decoder substitution is not free. GPT-2 decodes in $400.1$\,ms against the shallow decoder's
$184.3$, a factor of $2.2$, lifting end-to-end latency by $1.84\times$. Twelve transformer layers
replace two, and greedy decoding runs them once per output token. So the GPT-2 configuration is
slower on the axis the approach exists to optimise, less accurate than the encoder substitution
alone, and unstable in training. We report it as a negative result.

Decoding dominates inference in every configuration, at 74\% of the baseline's total, because
the greedy loop re-runs the decoder over the full visual-plus-text sequence at each of the 32
steps. Key--value caching would reduce this substantially for all three, and its absence is a
property of the original implementation that we preserve for comparability rather than a
deliberate choice.

\subsection{Qualitative results}

Table~\ref{tab:qual} shows each configuration at its own best epoch, together with the GPT-2
configuration at the end of training, on three validation clips.

\begin{table*}[t]
\centering\small
\begin{tabular}{@{}p{0.14\textwidth}p{0.80\textwidth}@{}}
\toprule
\multicolumn{2}{@{}l}{\emph{Ground truth}: a man repelling down a side of a mountain covered in snow} \\
\midrule
CLIP+BERT & a man is using a shovel to clean up a snow covered area \\
SigLIP2+BERT & a man is using a shovel to pull a large snow covered mountain \\
SigLIP2+GPT-2 (ep.~2) & A man is shown using a rope to climb up a large rock \\
SigLIP2+GPT-2 (ep.~11) & A man is holding onto a rope and floating in the snow \\
\midrule
\multicolumn{2}{@{}l}{\emph{Ground truth}: A group of men are standing on a theater stage and preparing to do an acting scene} \\
\midrule
CLIP+BERT & a man is talking to a group of people in a room \\
SigLIP2+BERT & a man is talking to a group of people while talking to the camera \\
SigLIP2+GPT-2 (ep.~2) & A man is speaking into a microphone while a group of people are watching \\
SigLIP2+GPT-2 (ep.~11) & A man is standing and signing what a woman is saying in a recording \\
\midrule
\multicolumn{2}{@{}l}{\emph{Ground truth}: A Champion eyeglasses commercial showing different color frames} \\
\midrule
CLIP+BERT & a woman is talking about a video of a video \\
SigLIP2+BERT & a woman is talking about a video and then a man talks about it \\
SigLIP2+GPT-2 (ep.~2) & A woman is talking about a pair of sunglasses that are on a woman's face \\
SigLIP2+GPT-2 (ep.~11) & A woman is standing with a man and woman and people are talking and looking at them \\
\bottomrule
\end{tabular}
\caption{Generated captions, each configuration at its best epoch, with the GPT-2 configuration
also shown after overfitting. Detokenisation artefacts are removed for readability.}
\label{tab:qual}
\end{table*}

Two patterns are visible. First, the encoder substitution sharpens visual detail: SigLIP2
recovers ``mountain'' where the reproduction says only ``area'', consistent with its higher
scores across all four metrics.

Second, and less comfortably for our headline number, the GPT-2 configuration often produces the
most fluent and sometimes the best-grounded caption despite scoring lower. On the third clip it
is the only configuration to identify eyewear at all, and on the first the only one to recover
the rope. Its phrasing is more varied and more natural, which is precisely what $n$-gram metrics
penalise: CIDEr and BLEU reward overlap with reference wording, so a caption that describes the
same content in different words scores worse than a generic one that happens to reuse common
phrasing. We report this because it qualifies the comparison rather than because it overturns
it (the GPT-2 configuration remains slower, unstable, and unable to train to convergence at
this data scale), but a purely metric-driven reading would overstate the gap.

The two GPT-2 rows also illustrate the overfitting directly. At epoch~2 the model describes a
rope and a climb; by epoch~11 the same clip becomes ``floating in the snow'', and the theatre
clip acquires a recording and a woman who are not present. Grounding degrades into confabulation
as the decoder shifts from using the visual prefix to reproducing training-set phrasing.

\section{Limitations}
\label{sec:limitations}

\textbf{Single seed.} All reported results come from a single training run per configuration. On
a training set of 4,999 clips, run-to-run variation arising from initialisation, data order and
per-GOP frame sampling is plausibly of the order of one to two CIDEr points. Differences smaller
than this should not be read as meaningful, and we do not read them so.

\textbf{Partially unseeded baseline.} The CoCap reproduction was trained without a fixed random
seed for its first eight epochs, following an interrupted run that was resumed from a
checkpoint; the remaining epochs and both substitution runs use seed 42 throughout. Seed
variation is unbiased noise rather than a systematic effect, so this does not favour any
configuration, but it does mean the reproduction baseline is not exactly reproducible.

\textbf{Reduced scale.} We train on roughly one fifth of VATEX and evaluate on a 1,000-clip
subset, at a reduced sampling budget (Table~\ref{tab:deviations}). Our conclusions concern the
relative ordering of the three configurations under these conditions, not absolute performance.

\textbf{Training duration.} Validation CIDEr plateaus before the end of training in every
configuration, though not equally cleanly. For the reproduction, per-epoch gains fall from
$+2.5$ at epoch 7 to $+1.2$, $+0.7$, $+0.1$ and $+0.2$ over the four epochs that follow. The
SigLIP2 configuration is noisier: it gains $+2.8$ as late as epoch 8, then settles within
0.6 CIDEr across its final three epochs. The GPT-2 configuration had converged and begun
overfitting by epoch 3. No configuration is therefore truncated mid-improvement, and the
comparison is not confounded by differing degrees of under-training. What residual
under-training there may be sits on the SigLIP2 side, which would make the $+4.48$ CIDEr
encoder gain a conservative estimate rather than an inflated one.

\section{Conclusion}
\label{sec:conclusion}

We set out to determine which of CoCap's two dated components limits its accuracy. The answer is
the vision encoder, and only the vision encoder. Replacing CLIP with SigLIP2 improves every
caption metric ($+4.48$ CIDEr, $+1.34$ BLEU-4, $+0.66$ METEOR, $+0.51$ ROUGE-L) at no measurable
inference cost, because the two encoders are architecturally matched. Replacing the shallow
decoder with GPT-2, the substitution one would expect to help more, instead makes the system
slower, less accurate than the encoder substitution alone, and unstable: it reaches its best
score after two epochs and then loses $11.3$ CIDEr over the following nine while its training
loss falls by a factor of $5.7$.

The mechanism is capacity relative to data. At 4,999 training clips a 124M-parameter pretrained
decoder memorises the training set long before it exhausts what the visual representation can
tell it, whereas a two-layer decoder trained from scratch cannot. At this scale the shallow
decoder's limited capacity is protective rather than limiting, and modernising it is
counterproductive. Whether that reverses at full-VATEX scale, or under stronger regularisation of
the pretrained decoder, is the obvious next question and one our compute budget did not permit.

For practitioners the practical recommendation is narrow but clear: in compressed-domain video
captioning at modest data scale, upgrade the visual encoder and leave the decoder alone.

\appendix
\section{Reproduction notes}
\label{sec:repro}

Reproducing CoCap on a current toolchain required several source-level fixes, none of which are
discoverable from its documentation. We record them because they are the difference between the
release running and not running, and because our reproduction result depends on them.

\textbf{The compressed-domain reader.} CoCap reads motion vectors and residuals through a
separate native extension that builds a patched FFmpeg 5.1. Two patches are needed on any modern
Linux toolchain. FFmpeg 5.1 fails to assemble under binutils 2.41 or newer, which rejects an
unclipped shift constant in the inline assembly of \texttt{libavcodec/x86/mathops.h}; FFmpeg
fixed this after 5.1 by masking the operand. Separately, the extension predates NumPy~2, which
tightened \texttt{PyArray\_GETPTR3} to require \texttt{PyArrayObject~*} where the source passes
\texttt{PyObject~*}.

\textbf{Silent failure on unreadable video.} The reader catches every exception and returns
all-zero tensors alongside a failure flag that the calling code discarded. An unreadable clip
therefore trained as a black video, indistinguishable downstream from real data. We propagate the
flag and count such failures explicitly; all results reported here were produced with zero
occurrences.

\textbf{Tied weights in the optimizer grouping.} GPT-2 ties its output head to its token
embedding, so \texttt{named\_modules} reaches that tensor under two paths while
\texttt{named\_parameters} reports it once. CoCap's parameter grouping walks the former and looks
up in the latter, which raises on any decoder with tied weights.

\textbf{Feature pre-extraction.} Because \texttt{unfold\_sentences} treats each caption as a
training sample, a twelve-epoch run over ten captions per clip parses every bitstream around 120
times. We extract the compressed-domain features once and cache them (3.4\,MB per clip, 20.2\,GB
for 6,000 clips, about eight minutes), which leaves the GPU rather than the CPU as the
bottleneck.

Code, configurations and the exact commit for every run are released at
\url{https://github.com/nepalashim/compressed-video-captioning-ablation-cocap-siglip2-gpt2}; all experiments in this paper were produced at commit
\texttt{2bf9d07}.

\bibliographystyle{plain}
\bibliography{references}

\end{document}